\pdfoutput=1

\documentclass{article}

\usepackage{microtype}
\usepackage{graphicx}
\usepackage{booktabs}

\usepackage[utf8]{inputenc}
\usepackage[T1]{fontenc}

\usepackage{hyperref}

\usepackage[main, final]{neurips_2026}

\usepackage{amsmath}
\usepackage{amssymb}
\usepackage{mathtools}
\usepackage{amsthm}
\usepackage{multirow}
\usepackage[table]{xcolor}
\usepackage{array}
\usepackage{makecell}
\usepackage{subfigure}
\usepackage{caption}
\usepackage{wrapfig}

\usepackage[capitalize,noabbrev]{cleveref}

\theoremstyle{plain}
\newtheorem{theorem}{Theorem}[section]
\newtheorem{proposition}[theorem]{Proposition}

\theoremstyle{definition}
\newtheorem{definition}[theorem]{Definition}
\newtheorem{assumption}[theorem]{Assumption}
\theoremstyle{remark}

\usepackage[disable,textsize=tiny]{todonotes}

\title{Natural Jammers: When Non-Robust Features Become Antagonistic}

\author{Jiaming Zhang$^{1}$ \quad
    Fuyao Zhang$^{1}$ \quad
    Pengjun Xie$^{2}$ \quad
    Wei Yang Bryan Lim$^{1}$ \\
    $^{1}$Nanyang Technological University \qquad
    $^{2}$Alibaba Group \\[2pt]
    Project page: \url{https://jiamingzhang.netlify.app/naturaljammers}}

\begin{document}

\maketitle

\begin{abstract}
Deep neural networks are known to rely on non-robust features that align poorly with human perception. Prior work has largely framed these features as opportunistic shortcuts that help standard models on clean data. We show instead that non-robust cues can be antagonistic: they can act as \textbf{Natural Jammers} that actively mislead standard models on clean, unmodified images. Using disagreement between a standard model and robust models as an analysis lens, we identify a Jammed Set on which robust models outperform standard models.
We provide causal evidence for \textit{Spectral--spatial Locking}: jamming arises only when high-frequency components are precisely aligned with specific spatial structures. Disrupting this lock through spectral filtering or small geometric shifts can restore correct predictions. Crucially, we uncover a \textit{Substitution Paradox}: while individual jammers are fragile to geometric shifts, the aggregate prevalence of jamming remains approximately stable and is redistributed rather than eliminated under such transformations. Our results reframe natural non-robustness as a failure field over the empirical model--data geometry: one that redistributes rather than disappears under small transformations.
\end{abstract}

\section{Introduction}
\label{sec:intro}
Deep neural networks (DNNs) achieve remarkable success on visual tasks, yet their reasoning differs fundamentally from human perception. Rather than relying solely on semantic content, DNNs are highly sensitive to non-robust features—subtle, high-frequency patterns that predict labels but remain imperceptible to humans \cite{ilyas2019adversarial, gyevnar2025ai, liu2025comprehensive}. The prevailing view treats these features as opportunistic shortcuts: brittle patterns that standard models exploit as a crutch to reach correct decisions \cite{geirhos2020shortcut, kauffmann2025explainable, eriksson2025can}. This framing assumes non-robust signals and semantic labels work synergistically.

\begin{figure}[t]
\centering
\includegraphics[width=0.95\linewidth,trim=55 44 58 42,clip]{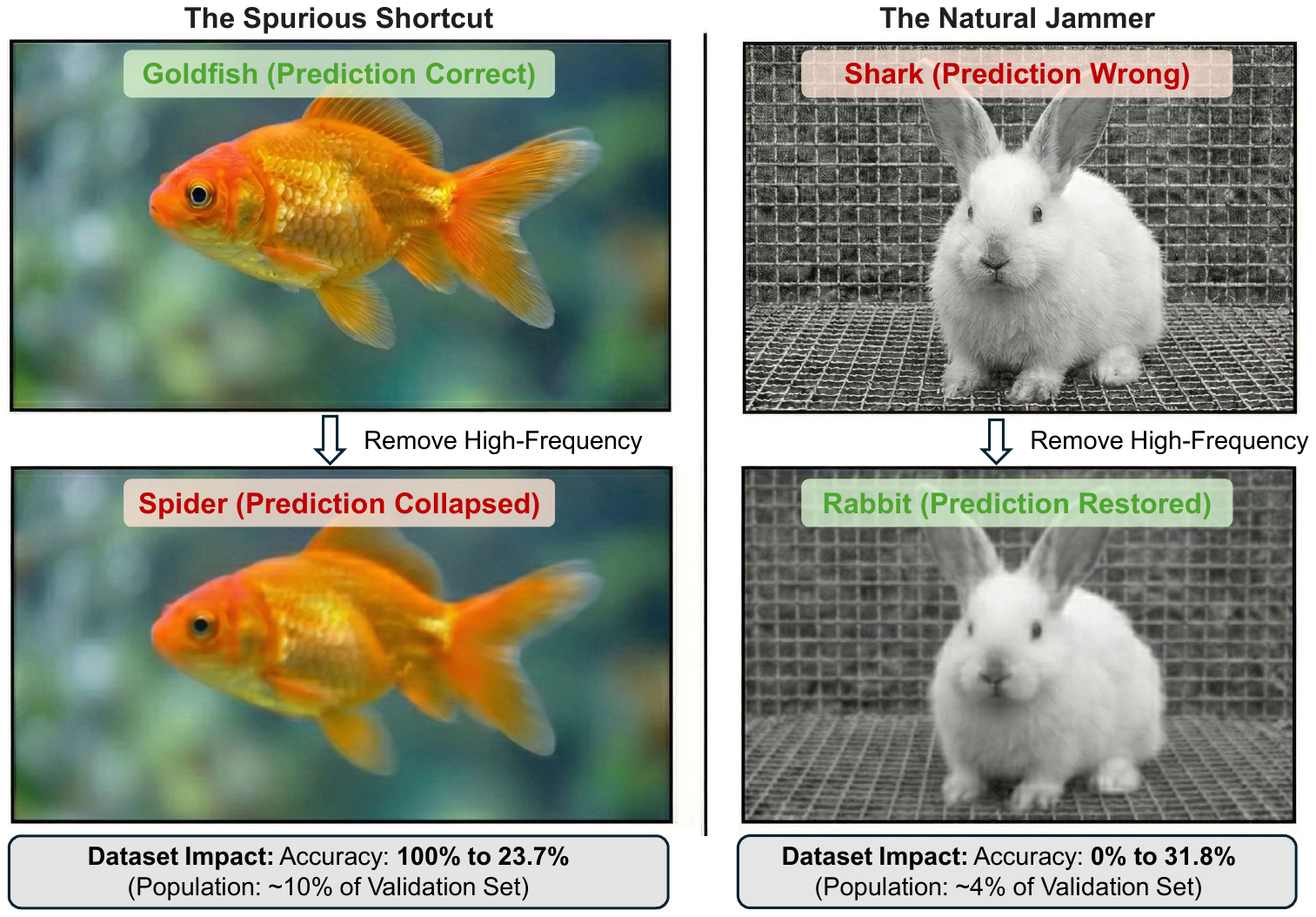}
\caption{\textbf{The Duality of Non-Robust Features.} (Left) \textbf{Spurious Shortcut} ($\mathcal{S}$): prediction relies on fragile non-robust cues and collapses when these cues are suppressed. (Right) \textbf{Natural Jammer} ($\mathcal{J}$): non-robust cues interfere with semantics-aligned evidence, suppressing them partially restores correct predictions. This illustrates that non-robust cues can be antagonistic rather than merely supportive. Accuracies are measured with the primary lens (standard vs.\ robust ResNet-50, $\ell_2$, $\epsilon=1$).}
\label{fig:1}
\end{figure}

We challenge this synergistic view by uncovering natural feature antagonism. As \cref{fig:1} illustrates, non-robust signals do not always aid classification—for a substantial fraction of natural images, they actively mislead standard inference. We term these signals \textbf{Natural Jammers}. This creates an empirical paradox: although the robustness-accuracy trade-off suggests standard models ($f_{\text{std}}$) should upper-bound clean performance \cite{tsipras2018robustness, suzuki2023adversarial, zhang2024improving, kafali2025dynamic, li2025triangular}, we identify a distinct \textbf{Jammed Set ($\mathcal{J}$)} where robustly trained models ($f_{\text{rob}}$) significantly outperform their standard counterparts on unmodified images. The standard model's reliance on non-robust features becomes a liability—it is misled by brittle cues that contradict the label, while the robust model often succeeds because it is less sensitive to such cues. We employ $f_{\text{rob}}$ as a diagnostic lens to disentangle robust semantic cues from non-robust correlations.

To explain this interference, we propose the \textbf{Spectral-Spatial Locking} mechanism through controlled experiments. Natural jammers arise from high-frequency patterns, but their disruptive influence depends critically on alignment with specific spatial configurations. This phase-locked coupling explains a striking duality: while these features mislead standard inference, they are also fragile. Breaking the lock—through spectral filtering (the Cleansing Paradox, Section~\ref{subsec:cleansing}) or minor geometric shifts (the Sensitivity Inversion, Section~\ref{subsec:xpattern})—renders the jammer inert, allowing the model's discriminative capacity to resurface.

This fragility raises a puzzle: if individual jammers are so easily disrupted, why does the model not achieve robust generalization? Our analysis reveals the \textbf{Substitution Paradox}: a 2\% edge crop removes about one third of the images in the Jammed Set, yet its total size remains nearly unchanged. Natural non-robustness is not a defect of particular images but a persistent property of the model--data geometry. When one jammer's spatial-spectral lock breaks, new jammers emerge elsewhere, keeping the overall vulnerability level nearly constant. Non-robustness is redistributed rather than eliminated by simple augmentations.

Our contributions are:
\begin{itemize}
\item \textbf{Feature Antagonism:} We formalize the Jammed Set ($\mathcal{J}$), showing that a significant portion of standard model errors stem from non-robust signals that contradict semantic labels.
\item \textbf{Spectral-Spatial Locking:} We provide causal evidence that non-robust features require precise alignment between high-frequency content and spatial structure to remain effective.
\item \textbf{Substitution Paradox:} We demonstrate that jamming is a stable, systemic property—individual non-robust features are fragile and interchangeable, following a redistribution pattern under geometric transformations.
\end{itemize}

\section{Landscape of Natural Non-Robustness}
\label{sec:landscape}

To move beyond a binary view of robust vs. non-robust features, we formalize a taxonomy based on the latent structure of the data distribution $\mathcal{D}$.

\begin{assumption}[Manifold Decomposition]
Following the latent manifold hypothesis \cite{ilyas2019adversarial}, we assume the input space $\mathcal{X}$ maps to a latent representation $\mathcal{Z}$ that can be decomposed into two subspaces, $\mathcal{Z} = \mathcal{Z}_r \times \mathcal{Z}_{nr}$, where $\mathcal{Z}_r$ and $\mathcal{Z}_{nr}$ denote the robust and non-robust manifolds, respectively.
\end{assumption}

The interaction between these manifolds determines whether a model's reliance on non-robust signals yields an opportunistic shortcut or a predictive failure. We utilize a standard classifier $f_{\text{std}}$ and an adversarially trained counterpart $f_{\text{rob}}$ to analyze these divergent behaviors.

\newcommand{\dcell}[2]{\begin{tabular}{@{}c@{}} #1 \\ \color{gray}\scriptsize (#2) \end{tabular}}

\begin{table}[t]
\caption{\textbf{Distribution of Feature Pathologies on ImageNet.} Data is reported for $\epsilon=1.0$ constraints. The proportions of $\mathcal{J}$ and $\mathcal{S}$ remain relatively stable across ResNet and WideResNet architectures, suggesting that jamming is not confined to a single architecture.}
\label{tab:representative_audit}
\begin{center}
\begin{small}
\begin{sc}
    \begin{tabular}{lcccc}
    \toprule
    Model & $\mathcal{J}$ (Jammed) & $\mathcal{S}$ (Spurious) & $\mathcal{C}$ (Common) & $\mathcal{G}$ (Uninformative) \\
    \midrule
    ResNet-18 ($\ell_2$) & \dcell{2,448}{4.90\%} & \dcell{6,168}{12.34\%} & \dcell{28,712}{57.42\%} & \dcell{12,672}{25.34\%} \\
    ResNet-50 ($\ell_2$) & \dcell{2,140}{4.28\%} & \dcell{5,002}{10.00\%} & \dcell{33,071}{66.14\%} & \dcell{9,787}{19.57\%} \\
    WRN-50-2 ($\ell_2$)  & \dcell{2,167}{4.33\%} & \dcell{4,698}{~9.40\%} & \dcell{34,543}{69.09\%} & \dcell{8,592}{17.18\%} \\
    \midrule
    \addlinespace[0.3em]
    ResNet-18 ($\ell_\infty$) & \dcell{2,502}{5.00\%} & \dcell{5,659}{11.32\%} & \dcell{29,221}{58.44\%} & \dcell{12,618}{25.24\%} \\
    ResNet-50 ($\ell_\infty$) & \dcell{2,252}{4.50\%} & \dcell{4,306}{~8.61\%} & \dcell{33,767}{67.53\%} & \dcell{9,675}{19.35\%} \\
    WRN-50-2 ($\ell_\infty$)  & \dcell{2,232}{4.46\%} & \dcell{4,145}{~8.29\%} & \dcell{35,096}{70.19\%} & \dcell{8,527}{17.05\%} \\
    \bottomrule
    \end{tabular}
\end{sc}
\end{small}
\end{center}
\vskip -0.1in
\end{table}

\subsection{Formal Taxonomy: The Quad-Set Partition}

Given a sample $(x, y) \in \mathcal{D}$ and a chosen reference pair $(f_{\text{std}}, f_{\text{rob}})$, we partition samples into four disjoint sets based on their predictions on the original clean input $x$:

\begin{itemize}
    \item \textbf{Common Set ($\mathcal{C}$):} $\{x \mid f_{\text{std}}(x) = y \wedge f_{\text{rob}}(x) = y\}$. Samples where features in $\mathcal{Z}_r$ and $\mathcal{Z}_{nr}$ are synergistic.
    \item \textbf{Spurious Set ($\mathcal{S}$):} $\{x \mid f_{\text{std}}(x) = y \wedge f_{\text{rob}}(x) \neq y\}$. The classic shortcut regime where $f_{\text{std}}$ succeeds via brittle patterns in $\mathcal{Z}_{nr}$.
    \item \textbf{Jammed Set ($\mathcal{J}$):} $\{x \mid f_{\text{std}}(x) \neq y \wedge f_{\text{rob}}(x) = y\}$. Samples where $f_{\text{std}}$ is misguided by antagonistic signals in $\mathcal{Z}_{nr}$.
    \item \textbf{Uninformative Set ($\mathcal{G}$):} $\{x \mid f_{\text{std}}(x) \neq y \wedge f_{\text{rob}}(x) \neq y\}$. Samples where signals in both manifolds are insufficient for classification.
\end{itemize}

While $\mathcal{S}$ represents the well-documented reliance on human-irrelevant shortcuts, $\mathcal{J}$ exposes a more pathological interaction where non-robust features actively contradict the ground-truth logic:

\begin{definition}[Natural Jammer]
A \textbf{Natural Jammer} is a non-robust signal $\mathbf{z}_{nr} \in \mathcal{Z}_{nr}$ that exhibits feature antagonism, such that its predictive direction contradicts the semantic label $y$.

Operationally, for a standard model $f_{\text{std}}$, an input $x$ is said to be \textit{jammed} if there exists a semantics-preserving transformation $T \in \mathcal{T}$ (e.g., spectral filtering or geometric cropping) such that:
\begin{equation}
    f_{\text{std}}(x) \neq y \quad \text{and} \quad f_{\text{std}}(T(x)) = y.
\end{equation}
In this context, $T$ acts as a de-locking operator that suppresses the antagonistic non-robust signal, allowing the model's latent discriminative capacity to resurface.
\end{definition}

\subsection{Empirical Prevalence and Stability}

We conduct a representative audit on the ImageNet validation set \cite{russakovsky2015imagenet} ($N=50,000$) to evaluate the prevalence of these sets across different architectures. The membership of the set is defined on the original inputs and, therefore, on the unmodified images $\text{Acc}_{f_{\text{std}}}(\mathcal{S})=100\%$ and $\text{Acc}_{f_{\text{std}}}(\mathcal{J})=0\%$ by construction (and analogously for $f_{\text{rob}}$). As shown in \cref{tab:representative_audit}, the Jammed Set $\mathcal{J}$ consistently accounts for a significant portion of the error manifold.

The consistency of these proportions (4.3\%--5.0\% for $\mathcal{J}$) across architectures suggests that natural jamming is not confined to a single evaluated model. However, this statistical stability does not necessarily imply that the ``Jammed'' status is an intrinsic property of the image. If the effect stems from universal data pathologies, the samples in $\mathcal{J}$ should overlap significantly across different models. Conversely, if it is a manifestation of specific inductive biases, these sets should diverge. We investigate this question by dissecting the spectral and spatial dependencies of these features in the next section.

\section{Spectral-Spatial Locking: The Core Evidence}
\label{sec:core_evidence}

The divergent behaviors of $f_{\text{std}}$ and $f_{\text{rob}}$ on the Jammed and Spurious sets suggest that these models prioritize different structural components within the same image. In this section, we conduct a series of diagnostic experiments to isolate whether these non-robust features are defined by their frequency content, their spatial location, or a combination of both. Our results lead to the discovery of a \textbf{Spectral-Spatial Locking} mechanism.

\textbf{Evaluation convention.} All sets $\mathcal{C},\mathcal{S},\mathcal{J},\mathcal{G}$ are defined using predictions on the original clean inputs. We then apply transformations and evaluate predictions on the same fixed subset of samples, without re-computing set membership after transformation. Unless stated otherwise, all experiments in this section use the primary lens (standard vs.\ robust ResNet-50, $\ell_2$, $\epsilon=1$; checkpoints from \citet{salman2020adversarially}).

\begin{figure}[t]
\centering
  \includegraphics[width=0.49\linewidth]{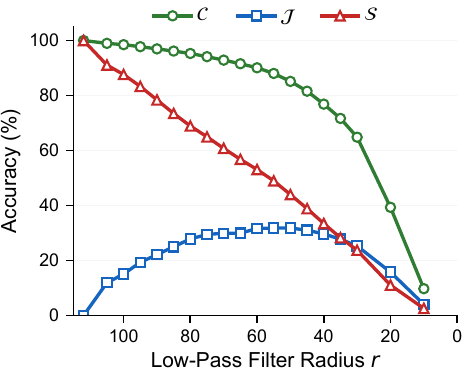}
  \hfill
  \includegraphics[width=0.49\linewidth]{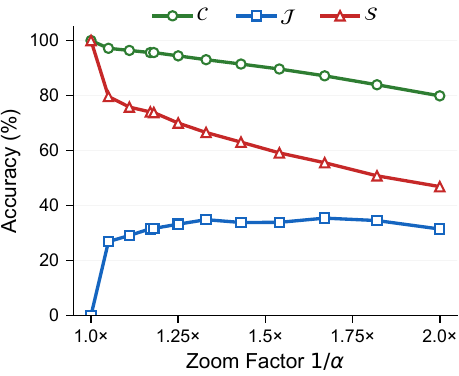}
\caption{\textbf{The Duality of Sensitivity.} (a) In the spectral domain, a \textit{Cleansing Paradox} is observed: removing high-frequencies restores accuracy for the Jammed Set ($\mathcal{J}$) while destroying it for the Spurious Set ($\mathcal{S}$). (b) In the spatial domain, zooming induces an \textit{Sensitivity Inversion} transition, showing the reversal of dominance between $\mathcal{J}$ and $\mathcal{S}$ features. All curves use the primary lens.}
\label{fig:sensitivity}
\end{figure}

\subsection{The Cleansing Paradox: Spectral Sensitivity}
\label{subsec:cleansing}

We begin by testing whether the interference in the Jammed Set is an intrinsic property of high-frequency (HF) patterns. We apply a low-pass filter (LPF) to progressively remove HF components and observe the accuracy of the standard model $f_{\text{std}}$.

\textbf{Protocol.} For each image $x$, we compute its 2D Fourier transform $\mathcal{F}(x)$, apply a circular hard-cutoff mask $M_r$ with varying radius $r \in [10, 112]$ (on a $224 \times 224$ spectrum), and reconstruct the filtered image $\tilde{x}_r = \mathcal{F}^{-1}(M_r \cdot \mathcal{F}(x))$. A smaller $r$ corresponds to more aggressive high-frequency removal.

\textbf{Results.} As shown in Figure~\ref{fig:sensitivity}(a), removing HF content yields opposite outcomes for the two sets: \textbf{Spurious Set ($\mathcal{S}$):} Accuracy collapses from 100.0\% to 23.7\% at $r=30$. This confirms that these predictions rely entirely on high-frequency shortcuts. \textbf{Jammed Set ($\mathcal{J}$):} Accuracy \textit{recovers} from 0.0\% to 31.8\% at $r=50$. \textbf{Common Set ($\mathcal{C}$):} Accuracy degrades gracefully from 100.0\% to 85.1\% at $r=50$, demonstrating that robust semantic features are largely preserved in the low-frequency skeleton. These findings are diagnostic evidence of the \textit{spectral fragility} of non-robust features: their predictive utility is tied to specific frequency supports (HF patterns).

\subsection{The Sensitivity Inversion: Spatial Localization}
\label{subsec:xpattern}

To determine where these jammers are located, we examine model sensitivity to spatial transformations.

\textbf{Protocol.} We apply a center-crop zoom $T_\alpha$, where the image is cropped to $\alpha \times 224$ pixels and resized back to $224 \times 224$ with bilinear interpolation. We sweep the ratio $\alpha$ from $1.0$ (original) to $0.5$ ($2\times$ zoom).

\textbf{Results.} Figure~\ref{fig:sensitivity}(b) reveals a distinct \textbf{Sensitivity Inversion}: as we zoom in, accuracy on \textbf{Jammed Set ($\mathcal{J}$)} increases while accuracy on \textbf{Spurious Set ($\mathcal{S}$)} decreases. Since zooming removes peripheral regions, this pattern is consistent with the hypothesis that many interfering cues are spatially localized near image boundaries or object context, where they can conflict with central semantic evidence. The accuracy on \textbf{Common Set ($\mathcal{C}$)} remains relatively stable, gracefully decreasing from 100\% to 79.9\%. Robust semantic features exhibit geometric invariance. This sensitivity highlights the \textit{spatial fragility} of the underlying non-robust signals. This recovery is not specific to the resampling method: it appears with all six interpolation kernels we tested and when a 2\% border is painted over instead of cropped.

\subsection{Frequency Grafting: A Causality Test}
\label{subsec:grafting}

The preceding experiments suggest a strong coupling between spectral content and spatial structure. To provide causal evidence, we test whether the interference in $\mathcal{J}$ is an intrinsic HF signal (analogous to a universal adversarial perturbation) or if its efficacy is strictly dependent on the host image's geometry. We perform a frequency grafting experiment that transplants the HF components of a Jammed image onto the LF skeleton of a Common image.

\textbf{Protocol.} Let $x_j \in \mathcal{J}$ be a Jammed image and $x_c \in \mathcal{C}$ be a Common image. We construct a hybrid image $x_{\text{hybrid}}$ by combining their spectral components as follows:
\begin{equation}
    x_{\text{hybrid}} = \mathcal{F}^{-1} \left( M_r \odot \mathcal{F}(x_c) + (1 - M_r) \odot \mathcal{F}(x_j) \right)
\end{equation}
where $\mathcal{F}$ denotes the 2D Discrete Fourier Transform, $\odot$ represents element-wise multiplication, and $M_r$ is a circular low-pass binary mask with radius $r=50$. This operation preserves the semantic skeleton of the Common image while grafting onto it the exact high-frequency textures of the Jammed image. We define \emph{transferred toxicity} as the fraction of random pairs $(x_j,x_c)$ for which the hybrid inherits the source jammer's predicted class, i.e., $f_{\text{std}}(x_{\text{hybrid}})=f_{\text{std}}(x_j)$.

\textbf{Results.} As shown in \cref{tab:grafting}, the results demonstrate a near-complete collapse of predictive influence upon transplantation. While suppressing high-frequency components yields a 31.8 percentage-point recovery on the native jammed inputs (Section~\ref{subsec:cleansing}), the transferred toxicity to a foreign host is only 0.17\% (averaged over five random pairings), whereas reassembling a jammer from its own spectral components preserves its jammed prediction in 99.9\% of cases.

\begin{table}[t]
\caption{\textbf{Frequency Grafting: Toxicity Transfer Test.} We graft the HF of Jammed images onto Common images. \textit{Native Toxicity} represents the accuracy gain when HF is removed from the source host. \textit{Transferred Toxicity} measures the rate at which the wrong label of the jammer infects a new semantic host.}
\label{tab:grafting}
\begin{center}
\begin{small}
\begin{sc}
\begin{tabular}{lcc}
\toprule
 Transfer & Native Toxicity & Transferred Toxicity \\
 Scenario & {\color{gray}\scriptsize(0\% $\to$ X\%)} & (on new host) \\
\midrule
$\mathcal{J} \to \mathcal{C}$ & 31.8\% {\color{gray}\scriptsize(0\%$\to$31.8\%)} & \textbf{0.17\%} \\
\bottomrule
\end{tabular}
\end{sc}
\end{small}
\end{center}
\vskip -0.1in
\end{table}

This result confirms that Natural Jammers are not independently ``toxic'' patterns. Their capacity to mislead the model is not an intrinsic property of the signal but emerges from a \textit{fragile resonance} with the host image's spatial structure. Once the spectral-spatial lock is broken by moving the HF signal to other host images, the feature becomes inefficient. This causal evidence shifts the understanding of non-robustness from a feature-level defect to a systemic interaction between spectrum and space.

\subsection{Formal Definition: Spectral--Spatial Locking} \label{subsec:formal_def}

The empirical evidence in the preceding sections reveals that natural non-robustness is neither a purely spectral defect nor a static spatial anomaly. Instead, it manifests as a structural dependency where the predictive utility of non-robust features is conditioned on the alignment of frequency and space. We formalize this mechanism as follows:
\begin{definition}[Spectral--Spatial Locking]
\label{def:locking}
Let $f$ be a classifier and $x \in \mathcal{X}$ denote an input sample. A non-robust predictive signal $\mathbf{z}$ is said to be \emph{spectral--spatially locked} if its contribution to the model's inference depends jointly on its frequency support and its spatial configuration. Formally, $\mathbf{z}$ is characterized by the following properties:
\begin{enumerate}
\item \textbf{Spectral sensitivity.} There exists a low-pass operator $\mathcal{L}_{\sigma}$ that suppresses frequencies above $\sigma$ such that removing these components changes the prediction, i.e., $f(\mathcal{L}_{\sigma}(x)) \neq f(x)$ for a non-trivial fraction of inputs exhibiting the signal.
\item \textbf{Spatial Fragility.} Let $\mathcal{T}$ be a group of semantics-preserving geometric transformations. There exists $T \in \mathcal{T}$ such that $f(\tau(x)) \neq f(x)$, demonstrating that the efficacy of $\mathbf{z}$ is sensitive to precise spatial alignment.
\item \textbf{Non-Transferability.} Let $x_A, x_B \in \mathcal{X}$ be two distinct samples. Grafting the spectral components of $x_A$ onto the spatial structure of $x_B$ fails to reproduce the original prediction bias, as the non-robust signal becomes functionally inert when the original lock is broken.
\end{enumerate}
\end{definition}
Definition~\ref{def:locking} establishes that individual jammers are fragile configurations that collapse when their spectral--spatial lock is disrupted. However, this fragility presents a paradox: if specific locks are so easily broken, why does the aggregate size of the Jammed Set remain approximately stable across models and augmentations? This indicates a redistribution of non-robustness across the data manifold. We will explore this in the next section.

\begin{table*}[t]
\caption{\textbf{Complete Cross-Model Landscape.} Analysis on ImageNet validation set ($N=50,000$). Within each architecture block, $f_{\text{std}}$ is fixed to that architecture's standard model; each row uses a different robust model $f_{\text{rob}}$ to define the quad-set partition on clean inputs.}
\label{tab:full_landscape}
\vskip 0.15in
\begin{center}
\begin{small}
\newcommand{\pcell}[2]{#1~{\color{gray}\scriptsize (#2)}}
\begin{tabular*}{\linewidth}{@{\extracolsep{\fill}} l cccc r @{}}
\toprule
\textbf{Configuration} & \textbf{Jammed} ($\mathcal{J}$) & \textbf{Spurious} ($\mathcal{S}$) & \textbf{Common} ($\mathcal{C}$) & \textbf{Uninformative} ($\mathcal{G}$) & \textbf{Rob. Acc.} \\
\midrule

\multicolumn{6}{l}{\textbf{ResNet-18} \hfill \textit{Standard Accuracy: 69.76\%}} \\
\midrule
$\ell_2, \epsilon=0.5$ & \pcell{2,639}{5.28\%} & \pcell{4,786}{9.57\%} & \pcell{30,094}{60.19\%} & \pcell{12,481}{24.96\%} & 65.5\% \\
$\ell_2, \epsilon=1.0$ & \pcell{2,448}{4.90\%} & \pcell{6,168}{12.34\%} & \pcell{28,712}{57.42\%} & \pcell{12,672}{25.34\%} & 62.3\% \\
$\ell_2, \epsilon=3.0$ & \pcell{1,924}{3.85\%} & \pcell{10,240}{20.48\%} & \pcell{24,640}{49.28\%} & \pcell{13,196}{26.39\%} & 53.1\% \\
\cmidrule[0.3pt](lr){1-6}
$\ell_\infty, \epsilon=1.0$ & \pcell{2,502}{5.00\%} & \pcell{5,659}{11.32\%} & \pcell{29,221}{58.44\%} & \pcell{12,618}{25.24\%} & 63.5\% \\
$\ell_\infty, \epsilon=4.0$ & \pcell{1,867}{3.73\%} & \pcell{10,497}{21.00\%} & \pcell{24,383}{48.77\%} & \pcell{13,253}{26.51\%} & 52.5\% \\
\midrule

\multicolumn{6}{l}{\textbf{ResNet-50} \hfill \textit{Standard Accuracy: 76.15\%}} \\
\midrule
$\ell_2, \epsilon=0.5$ & \pcell{2,317}{4.63\%} & \pcell{3,803}{7.61\%} & \pcell{34,270}{68.54\%} & \pcell{9,610}{19.22\%} & 73.2\% \\
$\ell_2, \epsilon=1.0$ & \pcell{2,140}{4.28\%} & \pcell{5,002}{10.00\%} & \pcell{33,071}{66.14\%} & \pcell{9,787}{19.57\%} & 70.4\% \\
$\ell_2, \epsilon=3.0$ & \pcell{1,807}{3.61\%} & \pcell{8,463}{16.93\%} & \pcell{29,610}{59.22\%} & \pcell{10,120}{20.24\%} & 62.8\% \\
\cmidrule[0.3pt](lr){1-6}
$\ell_\infty, \epsilon=1.0$ & \pcell{2,252}{4.50\%} & \pcell{4,306}{8.61\%} & \pcell{33,767}{67.53\%} & \pcell{9,675}{19.35\%} & 72.0\% \\
$\ell_\infty, \epsilon=4.0$ & \pcell{1,812}{3.62\%} & \pcell{7,948}{15.90\%} & \pcell{30,125}{60.25\%} & \pcell{10,115}{20.23\%} & 63.9\% \\
\midrule

\multicolumn{6}{l}{\textbf{WideResNet-50-2} \hfill \textit{Standard Accuracy: 78.48\%}} \\
\midrule
$\ell_2, \epsilon=0.5$ & \pcell{2,202}{4.40\%} & \pcell{3,887}{7.77\%} & \pcell{35,354}{70.71\%} & \pcell{8,557}{17.11\%} & 75.1\% \\
$\ell_2, \epsilon=1.0$ & \pcell{2,167}{4.33\%} & \pcell{4,698}{9.40\%} & \pcell{34,543}{69.09\%} & \pcell{8,592}{17.18\%} & 73.4\% \\
$\ell_2, \epsilon=3.0$ & \pcell{1,910}{3.82\%} & \pcell{7,701}{15.40\%} & \pcell{31,540}{63.08\%} & \pcell{8,849}{17.70\%} & 66.9\% \\
\cmidrule[0.3pt](lr){1-6}
$\ell_\infty, \epsilon=1.0$ & \pcell{2,232}{4.46\%} & \pcell{4,145}{8.29\%} & \pcell{35,096}{70.19\%} & \pcell{8,527}{17.05\%} & 74.7\% \\
$\ell_\infty, \epsilon=4.0$ & \pcell{2,000}{4.00\%} & \pcell{7,037}{14.07\%} & \pcell{32,204}{64.41\%} & \pcell{8,759}{17.52\%} & 68.4\% \\
\midrule

\multicolumn{6}{l}{\textbf{ViT-B/16} \hfill \textit{Standard Accuracy: 81.07\%}} \\
\midrule
$\ell_\infty, \epsilon=1.0$ & \pcell{1,569}{3.14\%} & \pcell{5,467}{10.93\%} & \pcell{35,067}{70.13\%} & \pcell{7,897}{15.79\%} & 73.3\% \\
\midrule

\multicolumn{6}{l}{\textbf{ViT-B/32} \hfill \textit{Standard Accuracy: 75.91\%}} \\
\midrule
$\ell_\infty, \epsilon=1.0$ & \pcell{1,828}{3.66\%} & \pcell{6,511}{13.02\%} & \pcell{31,443}{62.89\%} & \pcell{10,218}{20.44\%} & 66.5\% \\
\bottomrule
\end{tabular*}
\end{small}
\end{center}
\end{table*}

\section{Substitution Paradox: Systemic Redistribution of Natural Non-Robustness}
\label{sec:generalization}

Section~\ref{sec:core_evidence} established that many clean-image failures of standard models are consistent with \emph{spectral--spatial locking}: interfering cues are fragile under mild spectral or geometric perturbations. This fragility raises a natural question: if individual jammers are so easily disrupted, why does non-robust vulnerability persist at scale?

This section provides system-level evidence that natural non-robustness is not confined to a small set of ``bad images'' but is a persistent outcome of the model--data interaction. We focus on three questions: \textbf{Q1:} Is the prevalence of $\mathcal{J}$ and $\mathcal{S}$ stable across architectures and robustness recipes? \textbf{Q2:} Are jammed/spurious identities intrinsic to images, or model-dependent? \textbf{Q3:} Under tiny semantics-preserving transformations, do these pathologies disappear or get redistributed?

\begin{assumption}[Robust Models as Analysis Lenses]
\label{assump:lens}
Robust models are used as analysis lenses to induce the quad-set partition via disagreement with a standard model. We do not treat any single robust model as a ground-truth oracle; instead, we test whether conclusions persist across multiple robust oracles spanning architectures and robustness recipes.
\end{assumption}

\paragraph{Experimental Setup.}
\label{subsec:setup_generalization}

All analyses are conducted on ImageNet validation ($N=50{,}000$). For each architecture family, we use its corresponding standard model as $f_{\text{std}}$ and a collection of robust models as candidate lenses $f_{\text{rob}}$ that vary in threat model ($\ell_2$ or $\ell_\infty$) and budget $\epsilon$.
\emph{Within each architecture block}, $f_{\text{std}}$ is fixed to that architecture's standard model, and each robust model defines a quad-set partition
$\mathcal{C},\mathcal{S},\mathcal{J},\mathcal{G}$
according to Section~\ref{sec:landscape} on the original clean inputs.

\subsection{Prevalence Across Architectures}
\label{subsec:landscape_full}

We first establish a baseline fact that will serve as a foundation for the remainder of the section: natural jamming is not rare, nor specific to a single robustness recipe. Table~\ref{tab:full_landscape} reports the quad-set partition induced by 17 robust lenses spanning CNNs and ViTs.

Across a wide range of architectures and robustness recipes, the Jammed Set consistently occupies a few percent of ImageNet validation set , and the Spurious Set occupies a larger but still substantial portion (roughly $\sim$7--20\%, depending on $\epsilon$). Natural jamming is thus not an isolated artifact of a particular model or robustness configuration; across these lenses, $|\mathcal{J}|$ is relatively stable compared to $|\mathcal{S}|$.

\subsection{Model-Specific Vulnerabilities}
\label{subsec:iou_confidence}

We next test whether jammed is an intrinsic property of the image or depends on the model used to define the partition. For two induced Jammed sets $\mathcal{J}_A$ and $\mathcal{J}_B$, we quantify agreement via Intersection-over-Union (IoU):
\[
\mathrm{IoU}(\mathcal{J}_A,\mathcal{J}_B)=\frac{|\mathcal{J}_A\cap \mathcal{J}_B|}{|\mathcal{J}_A\cup \mathcal{J}_B|},
\]
and analogously for $\mathcal{S}$.

\begin{figure}[t]
\centering
\includegraphics[width=0.58\linewidth]{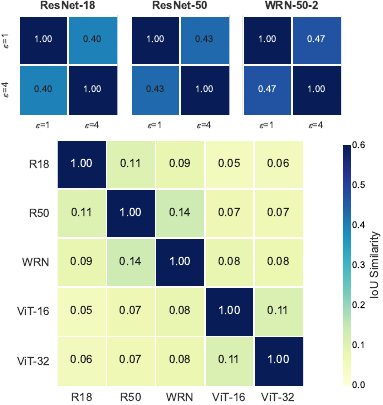}
\caption{\textbf{Jammed Set IoU Analysis ($\ell_\infty$).} \textit{Top row}: intra-architecture agreement when changing only robustness budget ($\epsilon=1$ vs $\epsilon=4$). \textit{Bottom}: cross-architecture IoU matrix. Agreement is modest within the same architecture and collapses across architectures, indicating that jammed identity is model-dependent.}
\label{fig:iou_heatmap}
\end{figure}

Agreement is modest within architecture and low across architectures. Figure~\ref{fig:iou_heatmap} shows that even within the same architecture, changing only the robustness budget yields moderate agreement (IoU $\approx 0.40$--$0.47$). Across architectures, IoU collapses (typically $<0.15$), and can be even smaller between CNNs and ViTs. These results indicate that \emph{which} images are jammed/spurious depends strongly on model inductive bias and robustness recipe, rather than being image-intrinsic.

While IoU captures \emph{identity}, confidence helps assess \emph{risk}: jammed errors are often high-confidence for standard models.
Table~\ref{tab:confidence_full} reports mean softmax confidence of $f_{\text{std}}$ and the corresponding robust lens $f_{\text{rob}}$ on the induced sets. Across CNNs, $f_{\text{std}}$ is typically \emph{more confident} than $f_{\text{rob}}$ on $\mathcal{J}$, consistent with a brittle but dominant cue overwhelming semantic evidence. ViTs show a reversal (robust confidence $>$ standard confidence), which may reflect architecture-specific interfering cues; however, the ViT and CNN lenses also differ in training recipe, so this comparison cannot separate the two factors.

\begin{table}[t]
\caption{\textbf{Confidence Comparison: Standard vs Robust Models.} Each cell reports $f_{\text{std}}$ / $f_{\text{rob}}$ mean confidence (max softmax on predicted class). ViT entries use $\ell_\infty$, $\epsilon$=1.0 only.}
\label{tab:confidence_full}
\begin{center}
\begin{footnotesize}
\begin{sc}
\setlength{\tabcolsep}{4pt}
\resizebox{\linewidth}{!}{
\begin{tabular}{@{}llcccccc@{}}
\toprule
& & \multicolumn{3}{c}{\textbf{Jammed} ($\mathcal{J}$)} & \multicolumn{3}{c}{\textbf{Spurious} ($\mathcal{S}$)} \\
\cmidrule(lr){3-5}\cmidrule(lr){6-8}
\textbf{Config} & $\boldsymbol{\epsilon}$ & \textbf{RN-18} & \textbf{RN-50} & \textbf{WRN-50-2} & \textbf{RN-18} & \textbf{RN-50} & \textbf{WRN-50-2} \\
\midrule
$\ell_2$ & 0.5 & .457\,/\,.440 & .511\,/\,.509 & .545\,/\,.556 & .571\,/\,.349 & .624\,/\,.412 & .684\,/\,.459 \\
$\ell_2$ & 1.0 & .470\,/\,.402 & .517\,/\,.464 & .555\,/\,.531 & .601\,/\,.303 & .661\,/\,.368 & .710\,/\,.427 \\
$\ell_2$ & 3.0 & .491\,/\,.318 & .544\,/\,.384 & .574\,/\,.449 & \textbf{.678}\,/\,.217 & \textbf{.726}\,/\,.267 & \textbf{.770}\,/\,.323 \\
$\ell_\infty$ & 1.0 & .468\,/\,.418 & .513\,/\,.485 & .554\,/\,.550 & .588\,/\,.316 & .643\,/\,.382 & .698\,/\,.445 \\
$\ell_\infty$ & 4.0 & .490\,/\,.319 & .538\,/\,.394 & .577\,/\,.464 & \textbf{.680}\,/\,.206 & \textbf{.717}\,/\,.270 & \textbf{.760}\,/\,.321 \\
\midrule
\multicolumn{2}{l}{ViT-B/16 \scriptsize($\ell_\infty$=1.0)} & \multicolumn{3}{c}{.504\,/\,\textbf{.701}} & \multicolumn{3}{c}{.640\,/\,.584} \\
\multicolumn{2}{l}{ViT-B/32 \scriptsize($\ell_\infty$=1.0)} & \multicolumn{3}{c}{.447\,/\,\textbf{.670}} & \multicolumn{3}{c}{.591\,/\,.541} \\
\bottomrule
\end{tabular}
}
\end{sc}
\end{footnotesize}
\end{center}
\vskip -0.1in
\end{table}

\subsection{The Substitution Paradox}
\label{subsec:structural}

Section~\ref{sec:core_evidence} demonstrated that individual jammed behaviors can be neutralized by mild spectral filtering or small geometric shifts, suggesting that jamming depends on fragile spectral--spatial alignment. We now formalize and test a stronger claim: \emph{tiny transformations can break many specific jammed identities, yet the overall prevalence of jamming remains approximately conserved}.

\begin{definition}[Transformed Partition and Substitution Statistics]
\label{def:substitution_stats}
Fix an architecture and its standard model $f_{\text{std}}$, together with a robust lens $f_{\text{rob}}$ that induces the original sets
$\mathcal{C},\mathcal{S},\mathcal{J},\mathcal{G}$ on clean inputs.

For a transformation $T$, define the transformed Jammed set by evaluating both models on $T(x)$:
\[
\mathcal{J}_T := \{x : f_{\text{std}}(T(x))\neq y \ \wedge\ f_{\text{rob}}(T(x))=y\},
\]
and analogously $\mathcal{S}_T,\mathcal{C}_T,\mathcal{G}_T$.
We quantify (i) \emph{identity stability} via $\mathrm{IoU}(\mathcal{J},\mathcal{J}_T)$, and (ii) \emph{prevalence stability} via the transformed set sizes $|\mathcal{J}_T|$ and $|\mathcal{S}_T|$.
\end{definition}

\begin{proposition}[Substitution Paradox (Empirical)]
\label{prop:substitution}
For small semantics-preserving geometric transformations $T$ (e.g., $\leq 8\%$ edge crop with resize), we observe simultaneously:
\begin{enumerate}
    \item \textbf{Identity instability:} $\mathrm{IoU}(\mathcal{J},\mathcal{J}_T)$ (and $\mathrm{IoU}(\mathcal{S},\mathcal{S}_T)$) is low.
    \item \textbf{Prevalence stability:} the set sizes $|\mathcal{J}_T|$ and $|\mathcal{S}_T|$ remain approximately unchanged relative to $|\mathcal{J}|$ and $|\mathcal{S}|$.
\end{enumerate}
Thus, small geometric changes tend to \textbf{redistribute} jammed/spurious identities across samples rather than eliminate non-robust vulnerability.
\end{proposition}

We apply minimal center crops removing only 2\%, 4\%, 6\%, or 8\% of edge pixels and resize back to $224\times224$. For each crop ratio, we compute the transformed partition (Definition~\ref{def:substitution_stats}) and report both set sizes and IoU against the original partition.

\begin{table}[t]
\caption{\textbf{Structural Fragility + Substitution Paradox.} Minimal edge crops substantially change the \emph{identity} of jammed/spurious samples (IoU drops sharply), yet the \emph{prevalence} (set size) remains approximately stable.}
\label{tab:structural}
\begin{center}
\begin{small}
\begin{sc}
\setlength{\tabcolsep}{5pt}
\renewcommand{\arraystretch}{1.15}
\begin{tabular}{@{}c@{\hskip 8pt}cc@{\hskip 10pt}cc@{}}
\toprule
& \multicolumn{2}{c}{\textbf{Set Size}} & \multicolumn{2}{c}{\textbf{IoU vs. Original}} \\
\cmidrule(lr){2-3} \cmidrule(lr){4-5}
\textbf{Crop} & $|\mathcal{J}_T|$ & $|\mathcal{S}_T|$ & $\mathcal{J}$ & $\mathcal{S}$ \\
\midrule
0\% (original) & 2,140 & 5,002 & 1.00 & 1.00 \\
\textbf{2\%} & 2,289 & 5,193 & \textbf{0.46} & 0.58 \\
4\% & 2,268 & 5,208 & 0.42 & 0.54 \\
6\% & 2,276 & 5,127 & 0.38 & 0.51 \\
8\% & 2,291 & 5,159 & 0.35 & 0.49 \\
\bottomrule
\end{tabular}
\end{sc}
\end{small}
\end{center}
\vskip -0.1in
\end{table}

\textbf{Identity instability (structural fragility):} Even a 2\% crop yields $\mathrm{IoU}(\mathcal{J},\mathcal{J}_T)=0.46$, implying that about one third of the original jammed images leave $\mathcal{J}$, and a similar number of new ones enter, under an almost imperceptible geometric perturbation. This supports the view that the underlying cues depend on fragile spatial alignment, consistent with spectral--spatial locking.

\textbf{Prevalence stability (substitution):} Despite the large change in identity, the transformed set sizes remain stable: $|\mathcal{J}_T|\approx 2.2$k and $|\mathcal{S}_T|\approx 5.1$k across crop ratios. This stability is approximate rather than exact: in a re-run that tracks individual images, a 2\% crop moves 738 of the 2,140 jammers out of $\mathcal{J}$ and 873 new images in. Thus, breaking many specific locks does not reduce overall vulnerability; instead, new jammed/spurious identities emerge elsewhere. This resolves the tension raised by Section~\ref{sec:core_evidence}: \emph{individual} jammers are fragile, yet \emph{aggregate} non-robustness persists as a systemic property and is redistributed under small transformations. Similar turnover appears under mild low-pass filtering and on six further datasets; our claim is limited to these tested mild transformations. In this view, natural non-robustness is a systemic \textbf{``tax''} that the model must pay to achieve high training accuracy.

We formalized and verified the \emph{Substitution Paradox}: tiny semantics-preserving geometric changes drastically alter jammed/spurious identities while leaving their prevalence approximately unchanged.

\section{Related Work}
\label{sec:related}

\subsection{Adversarial Robustness and Non-robust Features}

The vulnerability of deep neural networks to adversarial examples \cite{szegedy2013intriguing, goodfellow2014explaining, madry2017towards} has been increasingly understood through the lens of data-driven features. \citet{ilyas2019adversarial} famously posited that adversarial examples arise from \textit{non-robust features}---patterns that are highly predictive yet imperceptible to humans---and that these features are actively \emph{useful} for standard classification. This framework has led to a prevailing view of non-robust features as ``spurious shortcuts'' \cite{geirhos2020shortcut, shah2020pitfalls} that models exploit to achieve high standard accuracy, often resulting in an inherent trade-off between robustness and clean performance \cite{tsipras2018robustness, zhang2019theoretically, rebuffi2021fixing}.

This view has since been actively debated. \citet{kim2021distilling} show via information bottleneck that non-robust features are decomposable, with only a subset driving adversarial predictions. \citet{li2023adversarial} argue that non-robust features are paradigm-specific shortcuts useful only under supervised objectives, not across learning paradigms. Crucially, however, both works study features under \emph{intentional adversarial perturbation}---inputs explicitly crafted to expose model vulnerabilities.

We ask a fundamentally different question: can non-robust features mislead standard models on \emph{clean, unmodified natural images}, without any perturbation? We identify a Jammed Set ($\mathcal{J}$) where robust models outperform standard models on standard test images, and provide causal evidence for Spectral--Spatial Locking as the driving mechanism. The antagonism we document is not injected by an attacker but arises spontaneously from the interaction between standard training and the spectral-spatial structure of natural data. Our Substitution Paradox further reveals this to be a redistributing property of the model--data geometry: a model pair defines a failure field $j(x)=\mathbf{1}[f_{\text{std}}(x)\neq y \wedge f_{\text{rob}}(x)=y]$ over natural images, which are commonly assumed to lie near a low-dimensional data manifold \cite{fefferman2016testing,stutz2019disentangling}, and small transformations move the support of $j$ while approximately preserving its mass. This provides a complementary framework to studies of natural adversarial or hard examples \cite{hendrycks2021natural, recht2019imagenet, li2024naturalbench, liu2025hard, zhang2025beyond}, which rely on curated atypical subsets rather than identifying a structural mechanism within standard uncurated data.

\subsection{Spectral and Geometric Bias in Generalization}

The inductive biases of CNNs are often characterized by their sensitivity to specific input frequencies and spatial structures. From a Fourier perspective, models frequently exhibit a \textit{spectral bias}, relying on high-frequency components that are critical for generalization but lack robustness \cite{yin2019fourier, wang2020high, fang2024addressing, krizhevsky2012imagenet, xia2024waveattack}. Simultaneously, the lack of geometric invariance---often caused by aliasing in downsampling layers \cite{zhang2019making}---renders models fragile to even minor image transformations \cite{azulay2019deep, michaeli2025alias}. This fragility is often linked to a "texture bias," where models prioritize local patterns over global shapes \cite{geirhos2018imagenet, hermann2020origins}.

Our work synthesizes these views through the mechanism of Spectral--spatial Locking, providing causal evidence that natural jamming is contingent on the precise alignment of frequency and space. Furthermore, we address the stability of these vulnerabilities, i.e., the Substitution Paradox where the aggregate rate of jamming remains stable across the data manifold. This observation relates to the broader problem of underspecification \cite{d2022underspecification}, where models with identical training objectives exhibit divergent and unpredictable failure modes under distribution shifts \cite{taori2020measuring, arjovsky2019invariant, sagawa2019distributionally}.

\section{Conclusion}
\label{sec:conclusion}

We show that non-robust features on \emph{clean} natural images can be antagonistic: on a Jammed Set, robust models outperform standard models without modifying inputs. Using a standard-vs-robust partition of ImageNet-Val, we characterize Jammed and Spurious regimes and provide evidence for \emph{spectral--spatial locking} via frequency filtering, geometric perturbations, and frequency grafting.

At the system level, jammed identities are highly model-dependent while prevalence remains stable, yielding the \emph{Substitution Paradox}: small semantics-preserving transformations reshuffle jammed membership without reducing its overall rate.

\begin{ack}
This research / project is supported by A*STAR under its Japan-Singapore Joint Call: JST-A*STAR 2024 (Project ID: R2416IR139). This research is also supported by the Ministry of Education, Singapore, under its Academic Research Fund Tier 2 (Award MOE-T2EP20125-0005). The authors declare no competing interests.
\end{ack}

\bibliography{example_paper}
\bibliographystyle{abbrvnat}

\end{document}